\documentclass[runningheads]{llncs}
\usepackage[T1]{fontenc}
\usepackage{graphicx}
\usepackage{booktabs}
\usepackage{amsmath}
\usepackage{multirow}
\usepackage{float}
\usepackage{subfigure}
\usepackage{ulem}
\begin{document}
\title{TCMD: A Traditional Chinese Medicine QA Dataset for Evaluating Large Language Models}
\titlerunning{TCMD}
% \author{Anonymous Author\inst{1}}
% %
% \authorrunning{Anonymous Author}
% \institute{Anonymous Affiliation, Anonymous Address
% \email{anonymous.author@xxx.com}\\
% \url{http://anonymous.com}}

\author{Ping Yu\inst{1} \and
Kaitao Song\inst{1} \and
Fengchen He\inst{1} \and
Ming Chen\inst{2} \and Jianfeng Lu\inst{1}}
\authorrunning{P. Yu, K. Song et al.}
\institute{Nanjing University of
Science and Technology, Nanjing, 210094, Jiangsu, China 
\email{pingyu@njust.edu.cn}\\
\and
Nanjing University of Chinese Medicine, Nanjing, 210023, Jiangsu, China}

\maketitle              % typeset the header of the contribution
\begin{abstract}
The recently unprecedented advancements in Large Language Models (LLMs) have propelled the medical community by establishing advanced medical-domain models. However, due to the limited collection of medical datasets, there are only a few comprehensive benchmarks available to gauge progress in this area. In this paper, we introduce a new medical question-answering (QA) dataset that contains massive manual instruction for solving Traditional Chinese Medicine examination tasks, called TCMD. 
%Traditional Chinese Medicine (TCM) is a medical discipline derived from China,  which is renowned worldwide as a prominent traditional medical practice. The recently unprecedented advancements in Large Language Models (LLMs) have propelled the medical community to establish multiple benchmarks for development. Although some comprehensive medical benchmarks have been proposed, none of them are designed to systemically evaluate the capability of LLMs in the TCM domain. To fill in the blank in this area, we built a new QA dataset called TCMD under the guidance of the Chinese National Medical Licensing Examination Manual Instruction. 
Specifically, our TCMD collects massive questions across diverse domains with their annotated medical subjects and thus supports us in comprehensively assessing the capability of LLMs in the TCM domain. Extensive evaluation of various general LLMs and medical-domain-specific LLMs is conducted. Moreover, we also analyze the robustness of current LLMs in solving TCM QA tasks by introducing randomness. The inconsistency of the experimental results also reveals the shortcomings of current LLMs in solving QA tasks. We also expect that our dataset can further facilitate the development of LLMs in the TCM area.

\end{abstract}

\section{Introduction}

In recent years, medical research has emerged as a frontier field at the intersection of science, medicine, and technological innovation, employing state-of-the-art technologies to extend our understanding of health and disease. 
% Medical research stands as a frontier topic at the intersection of science, medicine, and innovation, which condense enormous advanced technologies to explore the limit of the boundary. 
Recently, driven by the success of large language models (LLMs) ~\cite{du2022glm,qwen,2023internlm,touvron2023llama}, building LLM-based medical applications~\cite{singhal2023towards,wang2023can,wang2023huatuo} has drawn enormous concerns. 
To evaluate these models and applications, various medical datasets~\cite{he2019applying,zhang2021cblue,wang2023cmb} are introduced, which promotes the prosperity and development of AI in healthcare.

Traditional Chinese Medicine (TCM) remains a crucial and fascinating field in Chinese medicine. So, to advance the integration of TCM with LLMs, researchers have collected or generated extensive TCM corpora and trained LLMs specifically for TCM~\cite{zhu2023ShenNong,zhu2023ChatMed,yang2023zhongjing,wang2023huatuo}. Additionally, several Chinese LLM providers have incorporated TCM data into their training datasets~\cite{baichuan2023baichuan2,2023internlm} to ensure these models are well-aligned with TCM knowledge. However, few of the current datasets are designed to evaluate the abilities of LLMs in TCM as some of them do not include samples from TCM, while some treat samples from TCM as part of their comprehensive evaluation.
Another important point is that most LLMs or LLM-powered applications in the TCM domain rely on subjective criteria such as usability and smoothness to assess their alignment with human expectations. However, these subjective metrics are labour-intensive and often lack consistency. Therefore, there is a pressing need for a new dataset that enables objective evaluation to address this gap.

As~\cite{chang2023survey} suggested, it is a natural option to use problems from standard examinations to objectively assess medical LLMs. For example, ~\cite{li2021mlec,kung2023performance,sharma2023performance} propose to use questions from the United States Medical Licensing Examination (USMLE) to test the performance of LLMs. Some of the tested LLMs perform well on these datasets, and a few of them even surpass the passing line for humans.

Following this principle, we propose building a multiple-choice QA (MCQA) dataset called TCMD to evaluate LLMs' performance in TCM objectively. Inspired by ~\cite{liu2023benchmarking}and~\cite{wang2023cmb}, we collect multiple-choice problems related to the Chinese National Medical Licensing Examination For Traditional Chinese Medicine and their corresponding explanations. 
Subsequently, we filter and organize the questions according to the official examination manual to ensure comprehensive coverage of all subjects mentioned in the instructions. This process guarantees that the distribution of questions across various medical topics aligns with the guidelines.
Furthermore, we introduce two types of multiple-choice problems not included in the two benchmarks. 
With this new dataset, we benchmark the performance of general LLMs, common medical LLMs, and TCM-domain-specific LLMs. We also propose to evaluate the robustness of LLMs by introducing randomness in the test, i.e., testing LLMs with the question whose options are shuffled. Surprisingly, the consistency of the predicted answers is quite poor, even for LLMs with high scores in normal evaluation. More details of the result will be discussed in Section \ref{sec:exp}. 

Overall, the main contributions and findings of this paper can be summarized as:
\begin{enumerate}
    \item We build a new dataset to evaluate the capability of different LLMs in solving TCM tasks. It contains 4 types of multiple-choice questions and their explanations in the style of the Chinese National Medical Licensing Examination. The questions are filtered and organized under the guidance of the official manual instruction.
    \item We conducted extensive analysis to investigate different LLMs and found that the general LLMs perform better than medical and TCM LLMs on average. Moreover, the consistency of answers predicted by LLMs when facing questions with shuffled options is far from satisfying.
    \item Moreover, we investigate the ensemble of predictions for questions with shuffled options using a voting mechanism and discover that this ensemble strategy can enhance final performance under certain conditions.
\end{enumerate}

\section{Related Work}
\subsection{LLMs for Medical Application}
In recent years, there has been an increasing interest in the application of LLMs in the field of medicine. Researchers have explored various approaches to leverage LLMs for medical tasks such as biomedical information extraction~\cite{kartchner2023zero,wang2023exploring,yang2022large}, clinical text analysis~\cite{yang2022large,chervenak2023promise}, medical queries~\cite{zakka2023almanac,hashemi2023dense}, and so on.
Certain researchers take an additional stride by directly replacing the doctor with LLMs~\cite{xiong2023doctorglm,li2023chatdoctor}. Particularly, in a double-blind trial, MedGPT, proposed by Medlinker, demonstrated a medical concordance rate of 96\% with attending physicians from Three-A hospital\footnote{\url{https://www.medlinker.com/news/201}}. Also, Med-PALM 2~\cite{singhal2023towards} becomes the first LLM to pass the United States Medical Licensing Examination style questions in MultiMedQA~\cite{singhal2022large} with an expert-level score.
Through such a plethora of studies, we can envision that in the near future, LLMs will be a great help to health workers.

\subsection{Medical Evaluation Datasets for LLMs}
%谷歌的palm2
% Due to the high labelling cost attributed to the employment of medical expert annotators and the privacy, medical datasets are hard to construct. Besides, LLM evaluation requires well-designed samples with diverse formats and rich knowledge. So, it is hard to evaluate the medical abilities of LLMs.
Medical evaluation datasets for LLM are hard to construct as they require well-designed samples with diverse formats and rich medical knowledge.
He~\cite{he2019applying} proposes to collect medical questions from online health consultancy websites to construct medical QA datasets. The dataset is unsatisfying in terms of professionalism and fluency because the conversations on online platforms are usually not conventional.
To overcome the problem, Li~\cite{li2021mlec} proposes the first large-scale open-source Chinese medical MCQA dataset, i.e. MLEC-QA, sourced from the Chinese National Medical Licensing Examination. Liu~\cite{liu2023benchmarking} collects problems from the same source and performs a more detailed categorization, making it possible to measure in multiple dimensions. 
To better leverage the existing datasets, ~\cite{singhal2022large} collects several medical datasets covering medical questions from multiple countries to assess LLMs in medical question answering.

Most of these datasets collect as many problems as possible to evaluate medical abilities comprehensively. However, knowledge of different subjects holds varying significance in real-world applications. To address it, our dataset contains questions from every required medical subject in the national exam, and we also balance the distribution of questions of different subjects according to their official manual.

\section{Dataset Construction}
\subsection{Manual Instruction of CNMLE}
The manual instruction of CNMLE gives detailed information about the CNMLE, including question types, exam content, exam schedule, 
etc. It is curated by medical and educational experts to evaluate students' mastery of essential knowledge required for real-world medical practice. As a result, it can be the guiding principle to construct an evaluation dataset.

\paragraph{Question Types}According to the part for TCM, there are 4 types of questions:
\begin{itemize}
\item \textbf{A1}\textbackslash{}\textbf{A2}: a single multiple-choice problem with a statement or a clinical case.
\item \textbf{A3}: several questions with a shared stem consisting of a clinical history.
\item \textbf{B1}: several questions with shared options containing interrelated concepts.
\end{itemize}
For convenience, we set the number of sub-problems for A3 and B1 problems as 3 and 2. English examples for different types can be found in the dataset. 

\begin{figure*}[t]
    \centering
    \includegraphics[scale=0.6]{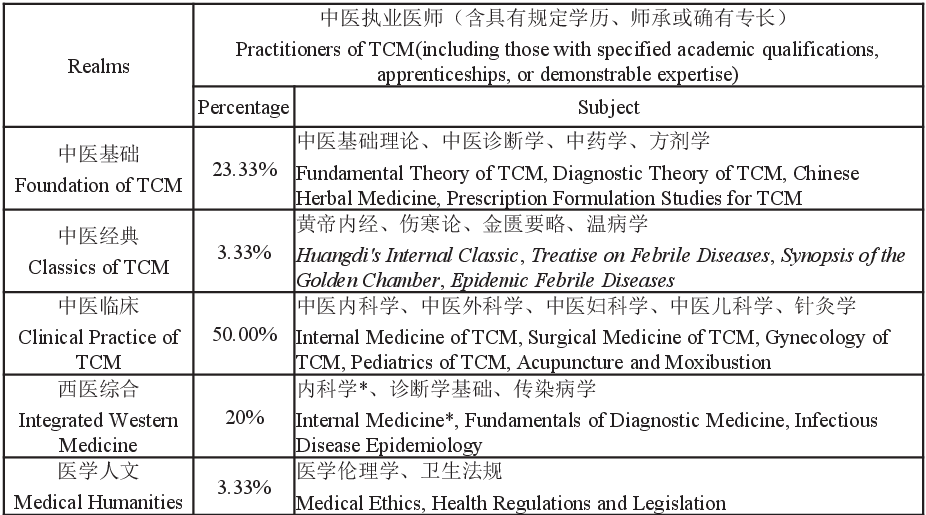}
    \caption{The detailed exam content of the CNMLE for TCM. The mark * indicates that the subject is not required for those with apprenticeships or
demonstrable expertise.}
    \label{fig:exam_content}
\end{figure*}

\paragraph{Exam Content} The instruction indicates that 5 medical realms and 18 subjects should be involved. A detailed list of them is shown in Figure \ref{fig:exam_content}, and so is the percentage of questions related to each realm. It's noticeable that the Clinical Practice of TCM accounts for half of the questions, which shows its importance in real-world diagnosis. The exam contains 600 questions in total.

\subsection{Data Collecting and Processing}
\paragraph{Problems Collection} 
Most of the questions are from two sources: reference books and TCM enthusiasts. Many of them are from actual examinations in different years. We also collect an explanation for each problem to consult the reasoning of LLMs.
% The acquisition sources for these questions primarily from two sources. Most examples in the test set are from questions in reference books which contains training tests for CNMLE. We also gather questions from an enthusiast in TCM to build the train set, and most of them are from real examinations in different years. Because some of the state-of-the-art methods require thought for solutions, we also collect an explanation for each problem.

\paragraph{Manual Verification and Processing} Most of these materials are in PDF format, so we first use optical character recognition to convert them to text files. Then, the files are verified by humans and programs to ensure their correctness. After verification, we de-duplicate the questions and group them based on which exam unit they belong to. Finally, medical experts are invited to check the examples.

\paragraph{Balance of Dataset}
To ensure that the dataset covers all the subjects and the test set follows the distribution from the manual instruction, we first invite some experts to label the subject of each collected problem. We drop or extend problems according to the gap between the real number and the target number of problems for different realms.

\paragraph{Data Statistics}
Finally, we collected 2851 questions for the train set and 600 questions for the test set. The number of questions of different types for different subjects is shown in Table \ref{tab:problem_statistic}.

\begin{table*}[!t]
\caption{The number of questions of different types for each subject grouped by realm. The first number indicates the train set, and the second is for the test set.
% For question types A3 and B1, the number represents the quantity of sub-questions.
}
\scriptsize
\centering
\begin{tabular}{l|c|c|c|c|c}
\toprule
\textbf{Subject} & \textbf{A1} & \textbf{A2} & \textbf{A3} & \textbf{B1} & \textbf{Total} \\
\midrule
Fundamental Theory of TCM & 259/25 & 3/0 & 0/0 & 87/10 & 349/35 \\
Diagnostic Theory of TCM & 197/17 & 136/6 & 45/0 & 131/12 & 509/35 \\
Chinese Herbal Medicine & 150/15 & 13/6 & 0/0 & 85/14 & 248/35 \\
Prescription Formulation Studies for TCM & 250/19 & 126/8 & 15/0 & 141/8 & 532/35 \\
\midrule
Huangdi's Internal Classic & 4/5 & 0/0 & 0/0 & 0/0 & 4/5 \\
Treatise on Febrile Diseases & 5/1 & 0/4 & 0/0 & 0/0 & 5/5 \\
Synopsis of the Golden Chamber & 4/1 & 0/4 & 0/0 & 0/0 & 4/5 \\
epidemic febrile diseases & 0/2 & 0/3 & 0/0 & 1/0 & 1/5 \\
\midrule
Internal Medicine of TCM & 4/18 & 0/27 & 0/27 & 8/18 & 12/90 \\
Surgical Medicine of TCM & 59/11 & 16/15 & 0/12 & 8/12 & 83/50 \\
Gynecology of TCM & 74/10 & 47/15 & 0/15 & 11/10 & 132/50 \\
Pediatrics of TCM & 54/10 & 41/15 & 0/15 & 3/10 & 98/50 \\
Acupuncture and Moxibustion & 114/12 & 60/18 & 0/18 & 37/12 & 211/60 \\
\midrule
Internal Medicine & 55/10 & 21/20 & 0/15 & 36/14 & 112/59 \\
Fundamentals of Diagnostic Medicine & 209/32 & 69/1 & 0/0 & 76/8 & 354/41 \\
Infectious Disease Epidemiology & 62/12 & 3/0 & 0/0 & 19/8 & 84/20 \\
\midrule
Medical Ethics & 35/6 & 0/0 & 0/0 & 19/4 & 54/10 \\
Health Regulations and Legislation & 45/6 & 1/0 & 0/0 & 13/4 & 59/10 \\
\bottomrule
\end{tabular}
\label{tab:problem_statistic}
\end{table*}

\begin{table*}[p]
\scriptsize
\caption{The performance of each model on the test set. The \textbf{FoT}, \textbf{CoT}, \textbf{CPoT}, \textbf{IWM}, \textbf{MH} in the header are short for Foundation of TCM, Classics of TCM, Clinical Practice of TCM, Integrated Western Medicine, Medical Humanities respectively. In the method column, "ICL" stands for In-Context Learning, and "COT" represents Chain-of-Thought. Models marked with an asterisk (*) are closed-source commercial models. A dash (-) in the score indicates that the model did not generate answers in the required format. 
The $\uparrow$ signifies the realm in which the model excels most out of the five domains, whereas $\downarrow$ indicates its weakest performance.}
\centering
\begin{tabular}{l|c|c|c|c|c|c|c|c}
\toprule
\multirow{2}{*}{\textbf{Model Name}} & \multirow{2}{*}{\textbf{Domain}} & \multirow{2}{*}{\textbf{Method}}  & \multicolumn{5}{c|}{\textbf{\underline{REALM}}} & \multirow{2}{*}{\textbf{Total}} \\
 & &  & \textbf{FoT} & \textbf{CoT} & \textbf{CPoT} & \textbf{IWM} & \textbf{MH} &  \\
\midrule

\multirow{2}{*}{ChatGPT4\textsuperscript{*}} & \multirow{2}{*}{General} & ICL & 0.550 & \textbf{0.350} & 0.563 & 0.667 & 0.850 &  0.583\\
 & & ICL + COT & 0.543 & \textbf{0.350} & 0.593 & 0.550 & 0.650 &  0.567\\
\midrule
\multirow{2}{*}{GPT-35-turbo-instruct\textsuperscript{*}} & \multirow{2}{*}{General} & ICL & 0.357 & \textbf{0.250} & 0.410 & 0.475 & 0.750 &  0.417\\
 & & ICL + COT & 0.350 & \textbf{0.000} & 0.310 & 0.250 & 0.550 &  0.305\\
\midrule
\multirow{2}{*}{Moonshot-v1-8k\textsuperscript{*}} & \multirow{2}{*}{General} & ICL & 0.693 & \textbf{0.500} & 0.713 & 0.708 & 0.600 &  0.697\\
 & & ICL + COT & 0.793 & \textbf{0.550} & 0.780 & 0.758 & 0.700 &  0.768\\
\midrule
\multirow{2}{*}{AquilaChat2-34B} & \multirow{2}{*}{General} & ICL & 0.421 & \textbf{0.300} & 0.407 & 0.483 & 0.650 &  0.430\\
 & & ICL + COT & 0.271 & \textbf{0.150} & 0.310 & 0.317 & 0.200 &  0.293\\
\midrule
\multirow{2}{*}{AquilaChat2-7B} & \multirow{2}{*}{General} & ICL & 0.293 & 0.350 & \textbf{0.287} & 0.375 & 0.350 &  0.310\\
 & & ICL + COT & 0.236 & \textbf{0.200} & 0.263 & 0.217 & 0.250 &  0.245\\
\midrule
\multirow{2}{*}{Baichuan2-13B-Base} & \multirow{2}{*}{General} & ICL & 0.657 & 0.600 & 0.607 & \textbf{0.592} & 0.650 &  0.617\\
 & & ICL + COT & 0.657 & \textbf{0.500} & 0.617 & 0.642 & 0.550 &  0.625\\
\midrule
\multirow{2}{*}{Baichuan2-7B-Base} & \multirow{2}{*}{General} & ICL & 0.536 & 0.400 & 0.547 & 0.617 & \textbf{0.350} &  0.547\\
 & & ICL + COT & 0.600 & 0.550 & \textbf{0.520} & 0.617 & 0.550 &  0.560\\
\midrule
\multirow{2}{*}{Baichuan-13B-Base} & \multirow{2}{*}{General} & ICL & 0.529 & \textbf{0.300} & 0.483 & 0.492 & 0.550 &  0.492\\
 & & ICL + COT & 0.486 & 0.450 & \textbf{0.410} & 0.458 & 0.750 &  0.450\\
\midrule
\multirow{2}{*}{Baichuan-7B} & \multirow{2}{*}{General} & ICL & 0.371 & \textbf{0.250} & 0.380 & 0.375 & 0.350 &  0.372\\
 & & ICL + COT & \textbf{0.314} & 0.450 & 0.317 & 0.442 & 0.400 &  0.348\\
\midrule
\multirow{2}{*}{ChatGLM3-6B} & \multirow{2}{*}{General} & ICL & 0.350 & \textbf{0.150} & 0.407 & 0.383 & 0.300 &  0.377\\
 & & ICL + COT & 0.464 & \textbf{0.250} & 0.420 & 0.400 & 0.300 &  0.417\\
\midrule
\multirow{2}{*}{ChatGLM2-6B} & \multirow{2}{*}{General} & ICL & 0.421 & \textbf{0.200} & 0.403 & 0.475 & 0.400 &  0.415\\
 & & ICL + COT & 0.450 & \textbf{0.200} & 0.480 & 0.542 & 0.550 &  0.478\\
\midrule
\multirow{2}{*}{ChatGLM-6B} & \multirow{2}{*}{General} & ICL & 0.321 & \textbf{0.250} & 0.250 & 0.350 & 0.400 &  0.292\\
 & & ICL + COT & 0.257 & 0.350 & 0.267 & 0.275 & \textbf{0.250} &  0.268\\
\midrule
\multirow{2}{*}{InternLM2-Chat-20B} & \multirow{2}{*}{General} & ICL & 0.614 & \textbf{0.400} & 0.653 & 0.650 & 0.600 &  0.633\\
 & & ICL + COT & 0.671 & \textbf{0.500} & 0.657 & 0.650 & 0.700 &  0.655\\
\midrule
\multirow{2}{*}{InternLM2-Chat-7B} & \multirow{2}{*}{General} & ICL & 0.557 & \textbf{0.300} & 0.660 & 0.500 & 0.450 &  0.585\\
 & & ICL + COT & 0.550 & \textbf{0.250} & 0.593 & 0.525 & 0.650 &  0.560\\
\midrule
\multirow{2}{*}{Qwen-14B-Chat} & \multirow{2}{*}{General} & ICL & 0.629 & \textbf{0.150} & 0.570 & 0.667 & 0.700 &  0.593\\
 & & ICL + COT & 0.629 & \textbf{0.350} & 0.563 & 0.675 & 0.700 &  0.598\\
\midrule
\multirow{2}{*}{Qwen-7B-Chat} & \multirow{2}{*}{General} & ICL & 0.400 & \textbf{0.200} & 0.443 & 0.467 & 0.450 &  0.430\\
 & & ICL + COT & \textbf{0.464} & 0.500 & 0.467 & 0.467 & 0.650 &  0.473\\
\midrule
\multirow{2}{*}{Vicuna-7B} & \multirow{2}{*}{General} & ICL & 0.264 & \textbf{0.050} & 0.227 & 0.292 & 0.200 &  0.242\\
 & & ICL + COT & 0.157 & \textbf{0.100} & 0.163 & 0.142 & 0.200 &  0.157\\
\midrule
\multirow{2}{*}{ChatMed-Consult} & \multirow{2}{*}{Medical} & ICL & 0.157 & 0.200 & 0.177 & \textbf{0.133} & 0.250 &  0.167\\
 & & ICL + COT & 0.050 & \textbf{0.000} & 0.030 & 0.008 & 0.150 &  0.033\\
\midrule
\multirow{2}{*}{ShenNong-TCM-LLM} & \multirow{2}{*}{TCM} & ICL & 0.129 & 0.300 & \textbf{0.100} & 0.100 & 0.300 &  0.120\\
 & & ICL + COT & 0.064 & 0.050 & \textbf{0.030} & 0.050 & 0.100 &  0.045\\
\midrule
\multirow{2}{*}{Huatuo-Llama-Med-Chinese} & \multirow{2}{*}{TCM} & ICL & 0.229 & 0.350 & 0.127 & \textbf{0.117} & 0.250 &  0.160\\

 & & ICL + COT &  - &  - &  - &  - &  - &  -\\

\bottomrule
\end{tabular}
\label{tab:main_result}
\end{table*}

\section{Experiments and Analysis} \label{sec:exp}
To make a thorough evaluation, we test various models on our datasets. The details of the experiment will be shown in this section.
\subsection{Experimental Setup}
\paragraph{Baseline}
We choose models of three different categories.
\begin{itemize}
    \item \textbf{General LLMs}: ChatGPT(ChatGPT4, GPT-35-turbo-instruct)~\cite{ChatGPT2023}, Kimi(Moonshot-v1-8k)~\footnote{\url{https://platform.moonshot.cn/}}, ChatGLM~\cite{du2022glm}, Baichuan~\cite{baichuan2023baichuan2}, QwenChat~\cite{qwen}, InternLMChat~\cite{2023internlm}, AquilaChat~\cite{2023Auila} and Vicuna~\cite{zheng2023judging}.
    \item \textbf{Common Medical LLMs}: ChatMed~\cite{zhu2023ChatMed}
    % , DoctorGLM~\cite{xiong2023doctorglm}
    \item \textbf{LLMs for TCM}: ShenNong-TCM-LLM\footnote{\url{https://github.com/michael-wzhu/ShenNong-TCM-LLM}}, Huatuo-Llama-Med-Chinese~\cite{wang2023huatuo}
\end{itemize}
Most models are open-source and can be found on the provided URL or Huggingface~\footnote{\url{https://huggingface.co/}}.

\paragraph{Prompt, Methods and Settings}
We hand-craft some prompts, while the others are from the model providers' evaluation code. These prompts can be found in the dataset. 
For prompting methods, we employ In-Context-Learning~\cite{brown2020language} and Chain-of-Thought~\cite{wei2022chain}. 
Specifically, we retrieve 3 examples for a demonstration from the train set using the Rank-BM25~\footnote{\url{https://github.com/dorianbrown/rank\_bm25}}. 
We also think splitting A3 and B1 questions into single sub-questions is against their purpose, so we feed the LLMs with the whole question.
As for the generation parameters, we use the default ones except that we set \textit{do\_sampe} to \textit{False} to avoid irreproducible results. We set \textit{max\_new\_tokens} to 128. For the same purpose, we set \textit{seed} to 0 for ChatGPT and Kimi.

\subsection{Main Result}
We use accuracy as our evaluation metric. The detailed scores for each model are listed in Table \ref{tab:main_result}.
Despite the fact that the medical and TCM LLMs used medical or TCM corpus during training, the overall performances of general LLMs are much better. We think the tuning process of medical LLMs and TCM LLMs only involves textual QA tasks, which may result in catastrophic forgetting~\cite{french1999catastrophic} of their abilities to answer multiple-choice questions. Among these LLMs, Moonshot-v1-8k using chain-of-thought earns the highest score. Among those open-source models, InternLM2-Chat-20B using chain-of-thought archives the best performance. Theoretically, if the testee answers correctly to over 60\% of the questions, he will pass the examination. So, Moonshot-v1-8k, Baichuan2-13b-Base and InternLM2-Chat-20B are the three passers of our evaluation. The training data of Baichuan2 and InternLM2 includes medical literature according to their technical report, and we think it contributes to their outstanding performance. Another noteworthy point is that ChatGPT4, one of the most powerful LLMs in the world, does not achieve a passing score, and the reason might be the lack of TCM knowledge in the training phase.

\subsection{Analysis}
We will analyze our results from the following four perspectives.
\paragraph{The performance in each realm}
Among all the 5 realms, most models perform terribly in Classics of TCM. We think the reason for the poor performance is that classical Chinese literature is not common in the training corpus but is ubiquitous in this realm. Although Classics of TCM only account for 3.33\% of the questions, it plays an important role in learning TCM, and the Chinese people have a strong passion for classical Chinese.
% So, we think the ability of LLMs to understand classical Chinese should be strengthened.

\paragraph{The performance on different question types}

Statistics about performance on different types are shown in Table \ref{tab:type_result}. To avoid redundancy, those results whose total accuracy is no less than 0.5 are analyzed. The statistics show that most of the LLMs perform poorly on type B1 and type A3. Type B1 questions usually contain confusing concepts, and some medical experts may even make mistakes about these concepts. Type A3 questions contain complex reasoning processes. So, the abilities of LLMs for concept discrimination and reasoning in TCM should be strengthened.

\paragraph{The birth time of models}
For each series of models, like ChatGLM and Baichuan, we can see that the newer versions of models are generally better than the old ones. Particularly, Baichuan2-7B-Base even beats Baichuan-13B-Base despite the huge gap in parameter size. The phenomenon proves that medical applications hold a significant strategic position among various LLM providers.

\paragraph{Is Chain-of-Thought a universal remedy?}
Chain-of-Thought (COT) is a powerful and well-validated reasoning method that can enhance the logical reasoning capabilities of LLMs. However, our experimental results show that using COT did not improve accuracy for several models. We believe this is because most models have not been trained on data from the TCM domain, leading to a lack of understanding of many TCM concepts. As a result, using COT may conflict with the models' internal knowledge, reducing their performance. Therefore, COT cannot be considered a universal solution for enhancing model performance in the TCM domain.

\begin{table*}[t]
\scriptsize
\caption{The performance of each model on different types of questions.
% The mark * indicates that the LLM with the method in the same line performs worst on this type of question.
}
\centering
\begin{tabular}{l|c|c|c|c|c}
\toprule
\textbf{Model} &\textbf{Method} & \textbf{A1} & \textbf{A2} & \textbf{A3} & \textbf{B1} \\
\midrule

\multirow{2}{*}{ChatGPT4\textsuperscript{*}} & ICL & 0.580 & 0.613 & \textbf{0.569} & 0.569\\
 & ICL + COT & 0.561 & \textbf{0.465} & 0.647 & 0.618\\
\midrule
\multirow{2}{*}{moonshot-v1-8k\textsuperscript{*}} & ICL & 0.741 & 0.746 & 0.686 & \textbf{0.590}\\
 & ICL + COT & 0.807 & 0.768 & \textbf{0.686} & 0.771\\
\midrule

\multirow{2}{*}{Baichuan2-13B-Base} & ICL & 0.651 & 0.725 & 0.588 & \textbf{0.479}\\
 & ICL + COT & 0.656 & 0.669 & \textbf{0.539} & 0.597\\
\midrule
\multirow{2}{*}{Baichuan2-7B-Base} & ICL & 0.599 & 0.662 & 0.520 & \textbf{0.375}\\
 & ICL + COT & 0.594 & 0.641 & 0.510 & \textbf{0.465}\\
\midrule
\multirow{2}{*}{InternLM2-Chat-20B} & ICL & 0.675 & 0.683 & 0.608 & \textbf{0.542}\\
 & ICL + COT & 0.646 & 0.662 & \textbf{0.627} & 0.681\\
\midrule
\multirow{2}{*}{InternLM2-Chat-7B} & ICL & 0.608 & 0.634 & 0.686 & \textbf{0.431}\\
 & ICL + COT & 0.580 & 0.585 & 0.569 & \textbf{0.500}\\
\midrule
\multirow{2}{*}{Qwen-14B-Chat} & ICL & 0.689 & 0.599 & 0.588 & \textbf{0.451}\\
 & ICL + COT & 0.679 & 0.683 & \textbf{0.412} & 0.528\\

\bottomrule
\end{tabular}

\label{tab:type_result}
\end{table*}

\begin{table*}[t]
\scriptsize
\caption{The statistics of accuracy and consistency for predictions to shuffled variants of the test set.}
\centering
\begin{tabular}{l|c|c|c|c|c|c}
\toprule
\textbf{Model} & \multicolumn{3}{c|}{\textbf{\underline{Accuracy}}} & \multicolumn{3}{c}{\textbf{\underline{Consistency}}} \\
& \textbf{Mean(STD)}&\textbf{Min} & \textbf{Max} & \textbf{Mean(STD)}&\textbf{Min} & \textbf{Max} \\
\midrule

Baichuan2-13B-Base & 0.593(0.016) & 0.560 & 0.633 & 0.767(0.224) & 0.200 & 1.000 \\
Baichuan2-7B-Base & 0.540(0.015) & 0.505 & 0.580 & 0.699(0.221) & 0.200 & 1.000 \\
Baichuan-13B-Base & 0.481(0.018) & 0.442 & 0.530 & 0.648(0.212) & 0.210 & 1.000 \\
ChatGLM3-6B & 0.385(0.018) & 0.335 & 0.437 & 0.604(0.225) & 0.200 & 1.000 \\
ChatGLM2-6B & 0.397(0.019) & 0.360 & 0.438 & 0.584(0.259) & 0.200 & 1.000 \\
Qwen-14B-Chat & 0.603(0.017) & 0.570 & 0.645 & 0.756(0.223) & 0.200 & 1.000 \\
Qwen-7B-Chat & 0.444(0.015) & 0.400 & 0.478 & 0.623(0.231) & 0.200 & 1.000 \\
InternLM2-Chat-7B & 0.592(0.012) & 0.570 & 0.628 & 0.795(0.220) & 0.200 & 1.000 \\

\bottomrule
\end{tabular}
\label{tab:shuffle_distribution}
\end{table*}

\begin{figure}[!t]
	\centering
    \includegraphics[scale=0.55]{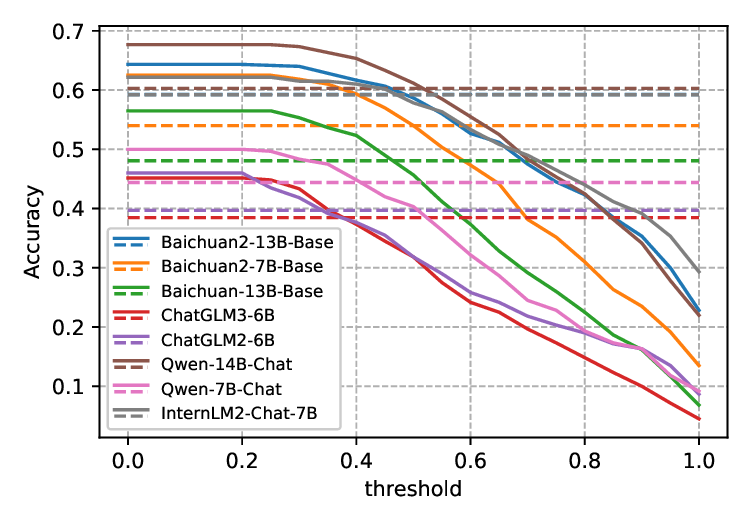}
	\caption{The accuracy of LLMs under different thresholds of consistency. The dashed line is the mean accuracy of the 120 variants from Table \ref{tab:shuffle_distribution}.}
    \label{fig:major_threshold}
\end{figure}

\subsection{Robustness Evaluation}
Some recent studies~\cite{berglund_reversal_2023} also indicate the LLM could be sensitive for the training order (e.g., a model trained on ``A is B'' cannot be generalized to the reverse pattern ``B is A''). Therefore, it is necessary to validate whether LLMs are robust in real-world applications. To this end, we also conduct experiments for robustness evaluation by using a simple strategy.

% Whether the LLMs are robust enough for real-world applications has always been a pain point for LLM assessment. The recent findings suggest that if a model is trained on a sentence of the form ``A is B'', it will not automatically generalize to the reverse pattern ``B is A''~\cite{berglund_reversal_2023}. We share the same worry with them, so we propose a simple way for robustness evaluation in this sub-section.

% cast a deep shadow on the problem. 

\subsubsection{Test Strategy}
To validate the robustness of LLMs, we review our preparation process for an exam during which we may exercise countless questions. To ensure that we grasp the question instead of simply memorizing it, we may shuffle the options and check the \textbf{consistency} of our answers. We consider this strategy to be useful for evaluating LLMs. So, we propose to \textbf{shuffle} the options and check whether the LLM gives a consistently good performance. To begin with, we define the consistency for question $q$ as
\begin{equation}
    \operatorname{cons}(q) = \frac{1}{n} \operatorname{max}(\operatorname{freq}(ans, pred), ans=A, B, C, D ,E),
\end{equation}
where $n$ is the number of variants for the question, $pred$ is the $n$ predictions made by the LLM and $\operatorname{freq}(ans, pred)$ gives the frequency of $ans$ among $pred$.

To minimize the influence of randomness, we collect the results of all possible mappers, i.e., the full permutations of the options. As we have 5 options for each question, we will have 120 shuffled variants for 1 question.
% Also, considering that the explanation contains the labels of options that are hard for automatic recognition and replacement, we only use ICL in the evaluation. 
Limited by resource, we only test 8 open-source well-performed models whose size is below 14B, which can be found in \ref{tab:shuffle_distribution} and the prompt method is ICL.

\subsubsection{Evaluation Result and Analysis}
We present and analyze the test results from the perspectives of accuracy and consistency. 

\paragraph{Accuracy} The statistical figures of models' accuracy across all shuffled variants of the test set are shown in Tabel \ref{tab:shuffle_distribution}. The gap between mean accuracy and the accuracy in Table \ref{tab:main_result} is narrow, and the standard diversion of accuracy remains small. These figures indicate that the overall performance of LLMs is stable.
% We can also observe that the maximum accuracy of Baichuan2-7B-Base exceeds the minimum accuracy of Baichuan2-13B-Base.
% , proving the possibility for LLMs to beat those with larger parameter sizes in a certain circumstance.

\paragraph{Consistency}\label{para:consistency}
We calculate the statistics of the consistency across the test set, which is shown in Table \ref{tab:shuffle_distribution}. As the figures indicate, even when facing two nearly identical questions, LLMs still have an unignorable chance to output different answers.

\subsubsection{Analysis on Ensemble  Methods}
% A common practice for dealing with possible inconsistent predictions is to integrate the results, i.e., ensemble. We integrate the results with 2 basic rules and 1 improved rule for further analysis.
A common and robust way to deal with predictions with uncertainty is to integrate multiple predictions, i.e., ensemble~\cite{dietterich2000ensemble}. We further investigate how the inconsistency of LLMs may affect this strategy by integrating the results. We take the most frequently occurring result among the 120 predicted answers as the final answer if the consistency passes the threshold. Otherwise, we consider the LLMs failed to generate the answer. The relationship between the threshold and the accuracy of the integrated result is shown in Figure \ref{fig:major_threshold}. For all models, the accuracy drops as the threshold rises. Particularly, the performance becomes unacceptable when the threshold goes high, which makes their robustness questionable. Interestingly, the integrated accuracy is higher than the mean accuracy from Table \ref{tab:shuffle_distribution}, which suggests that the integration may boost the performance if the threshold is very low.

\subsubsection{Summary}
Although the overall performance of most LLMs remains stable when tested with shuffled variants of questions, the consistency of all the LLMs is not so promising.
Fortunately, our experiments suggest that ensemble methods can alleviate this problem and may even boost the performance of LLMs. However, we should still think twice about adopting LLMs when faced with risk-averse tasks in complex and diverse environments, like surgery and financial systems.

\section{Conclusion}
In this paper, we propose a new dataset to systemically evaluate LLMs for their ability in TCM with MCQA questions under the guidance of the official examination manual instruction for CNMLE. The dataset contains questions from various realms and medical subjects. We also propose to evaluate the robustness of LLMs by the consistency of predictions when facing questions whose options are shuffled. We hope the dataset and our evaluation method will foster the advancement of LLMs in the field of TCM.

\bibliographystyle{splncs04}
\bibliography{tcmd_bib}

\begin{thebibliography}{10}
\providecommand{\url}[1]{\texttt{#1}}
\providecommand{\urlprefix}{URL }
\providecommand{\doi}[1]{https://doi.org/#1}

\bibitem{2023Auila}
BAAI: Aquilachat. \url{https://huggingface.co/BAAI} (2023)

\bibitem{baichuan2023baichuan2}
Baichuan: Baichuan 2: Open large-scale language models. arXiv preprint arXiv:2309.10305  (2023), \url{https://arxiv.org/abs/2309.10305}

\bibitem{berglund_reversal_2023}
Berglund, L., Tong, M., Kaufmann, M., Balesni, M., Stickland, A.C., Korbak, T., Evans, O.: The {Reversal} {Curse}: {LLMs} trained on "{A} is {B}" fail to learn "{B} is {A}" (Sep 2023), \url{http://arxiv.org/abs/2309.12288}, arXiv:2309.12288 [cs]

\bibitem{brown2020language}
Brown, T., Mann, B., Ryder, N., Subbiah, M., Kaplan, J.D., Dhariwal, P., Neelakantan, A., Shyam, P., Sastry, G., Askell, A., et~al.: Language models are few-shot learners. Advances in neural information processing systems  \textbf{33},  1877--1901 (2020)

\bibitem{chang2023survey}
Chang, Y., Wang, X., Wang, J., Wu, Y., Zhu, K., Chen, H., Yang, L., Yi, X., Wang, C., Wang, Y., et~al.: A survey on evaluation of large language models. arXiv preprint arXiv:2307.03109  (2023)

\bibitem{chervenak2023promise}
Chervenak, J., Lieman, H., Blanco-Breindel, M., Jindal, S.: The promise and peril of using a large language model to obtain clinical information: Chatgpt performs strongly as a fertility counseling tool with limitations. Fertility and Sterility  (2023)

\bibitem{dietterich2000ensemble}
Dietterich, T.G.: Ensemble methods in machine learning. In: International workshop on multiple classifier systems. pp. 1--15. Springer (2000)

\bibitem{du2022glm}
Du, Z., Qian, Y., Liu, X., Ding, M., Qiu, J., Yang, Z., Tang, J.: Glm: General language model pretraining with autoregressive blank infilling. In: Proceedings of the 60th Annual Meeting of the Association for Computational Linguistics (Volume 1: Long Papers). pp. 320--335 (2022)

\bibitem{french1999catastrophic}
French, R.M.: Catastrophic forgetting in connectionist networks. Trends in cognitive sciences  \textbf{3}(4),  128--135 (1999)

\bibitem{hashemi2023dense}
Hashemi, H., Zhuang, Y., Kothur, S.S.R., Prasad, S., Meij, E., Croft, W.B.: Dense retrieval adaptation using target domain description. In: Proceedings of the 2023 ACM SIGIR International Conference on Theory of Information Retrieval. pp. 95--104 (2023)

\bibitem{he2019applying}
He, J., Fu, M., Tu, M.: Applying deep matching networks to chinese medical question answering: a study and a dataset. BMC medical informatics and decision making  \textbf{19}(2),  91--100 (2019)

\bibitem{2023internlm}
InternLMTeam: Internlm: A multilingual language model with progressively enhanced capabilities. \url{https://github.com/InternLM/InternLM} (2023)

\bibitem{kartchner2023zero}
Kartchner, D., Ramalingam, S., Al-Hussaini, I., Kronick, O., Mitchell, C.: Zero-shot information extraction for clinical meta-analysis using large language models. In: The 22nd Workshop on Biomedical Natural Language Processing and BioNLP Shared Tasks. pp. 396--405 (2023)

\bibitem{kung2023performance}
Kung, T.H., Cheatham, M., Medenilla, A., Sillos, C., De~Leon, L., Elepa{\~n}o, C., Madriaga, M., Aggabao, R., Diaz-Candido, G., Maningo, J., et~al.: Performance of chatgpt on usmle: Potential for ai-assisted medical education using large language models. PLoS digital health  \textbf{2}(2),  e0000198 (2023)

\bibitem{li2021mlec}
Li, J., Zhong, S., Chen, K.: Mlec-qa: A chinese multi-choice biomedical question answering dataset. In: Proceedings of the 2021 Conference on Empirical Methods in Natural Language Processing. pp. 8862--8874 (2021)

\bibitem{li2023chatdoctor}
Li, Y., Li, Z., Zhang, K., Dan, R., Jiang, S., Zhang, Y.: Chatdoctor: A medical chat model fine-tuned on a large language model meta-ai (llama) using medical domain knowledge. Cureus  \textbf{15}(6) (2023)

\bibitem{liu2023benchmarking}
Liu, J., Zhou, P., Hua, Y., Chong, D., Tian, Z., Liu, A., Wang, H., You, C., Guo, Z., Zhu, L., et~al.: Benchmarking large language models on cmexam--a comprehensive chinese medical exam dataset. arXiv preprint arXiv:2306.03030  (2023)

\bibitem{ChatGPT2023}
OpenAI: Chatgpt: Language model. \url{https://www.openai.com/chatgpt} (2023)

\bibitem{sharma2023performance}
Sharma, P., Thapa, K., Dhakal, P., Upadhaya, M.D., Adhikari, S., Khanal, S.R.: Performance of chatgpt on usmle: Unlocking the potential of large language models for ai-assisted medical education. arXiv preprint arXiv:2307.00112  (2023)

\bibitem{singhal2022large}
Singhal, K., Azizi, S., Tu, T., Mahdavi, S.S., Wei, J., Chung, H.W., Scales, N., Tanwani, A., Cole-Lewis, H., Pfohl, S., et~al.: Large language models encode clinical knowledge. arXiv preprint arXiv:2212.13138  (2022)

\bibitem{singhal2023towards}
Singhal, K., Tu, T., Gottweis, J., Sayres, R., Wulczyn, E., Hou, L., Clark, K., Pfohl, S., Cole-Lewis, H., Neal, D., et~al.: Towards expert-level medical question answering with large language models. arXiv preprint arXiv:2305.09617  (2023)

\bibitem{qwen}
Team, Q.: Qwen technical report. arXiv preprint arXiv:2309.16609  (2023)

\bibitem{touvron2023llama}
Touvron, H., Martin, L., Stone, K., Albert, P., Almahairi, A., Babaei, Y., Bashlykov, N., Batra, S., Bhargava, P., Bhosale, S., et~al.: Llama 2: Open foundation and fine-tuned chat models. arXiv preprint arXiv:2307.09288  (2023)

\bibitem{wang2023huatuo}
Wang, H., Liu, C., Xi, N., Qiang, Z., Zhao, S., Qin, B., Liu, T.: Huatuo: Tuning llama model with chinese medical knowledge (2023)

\bibitem{wang2023exploring}
Wang, Q., Gao, Z., Xu, R.: Exploring the in-context learning ability of large language model for biomedical concept linking. arXiv preprint arXiv:2307.01137  (2023)

\bibitem{wang2023cmb}
Wang, X., Chen, G.H., Song, D., Zhang, Z., Chen, Z., Xiao, Q., Jiang, F., Li, J., Wan, X., Wang, B., et~al.: Cmb: A comprehensive medical benchmark in chinese. arXiv preprint arXiv:2308.08833  (2023)

\bibitem{wang2023can}
Wang, Z., Li, R., Dong, B., Wang, J., Li, X., Liu, N., Mao, C., Zhang, W., Dong, L., Gao, J., et~al.: Can llms like gpt-4 outperform traditional ai tools in dementia diagnosis? maybe, but not today. arXiv preprint arXiv:2306.01499  (2023)

\bibitem{wei2022chain}
Wei, J., Wang, X., Schuurmans, D., Bosma, M., Xia, F., Chi, E., Le, Q.V., Zhou, D., et~al.: Chain-of-thought prompting elicits reasoning in large language models. Advances in Neural Information Processing Systems  \textbf{35},  24824--24837 (2022)

\bibitem{zhu2023ShenNong}
Wei~Zhu, W.Y., Wang, X.: Shennong-tcm: A traditional chinese medicine large language model. \url{https://github.com/michael-wzhu/ShenNong-TCM-LLM} (2023)

\bibitem{xiong2023doctorglm}
Xiong, H., Wang, S., Zhu, Y., Zhao, Z., Liu, Y., Wang, Q., Shen, D.: Doctorglm: Fine-tuning your chinese doctor is not a herculean task. arXiv preprint arXiv:2304.01097  (2023)

\bibitem{yang2023zhongjing}
Yang, S., Zhao, H., Zhu, S., Zhou, G., Xu, H., Jia, Y., Zan, H.: Zhongjing: Enhancing the chinese medical capabilities of large language model through expert feedback and real-world multi-turn dialogue. arXiv preprint arXiv:2308.03549  (2023)

\bibitem{yang2022large}
Yang, X., Chen, A., PourNejatian, N., Shin, H.C., Smith, K.E., Parisien, C., Compas, C., Martin, C., Costa, A.B., Flores, M.G., et~al.: A large language model for electronic health records. NPJ Digital Medicine  \textbf{5}(1), ~194 (2022)

\bibitem{zakka2023almanac}
Zakka, C., Chaurasia, A., Shad, R., Hiesinger, W.: Almanac: Knowledge-grounded language models for clinical medicine. arXiv preprint arXiv:2303.01229  (2023)

\bibitem{zhang2021cblue}
Zhang, N., Chen, M., Bi, Z., Liang, X., Li, L., Shang, X., Yin, K., Tan, C., Xu, J., Huang, F., et~al.: Cblue: A chinese biomedical language understanding evaluation benchmark. arXiv preprint arXiv:2106.08087  (2021)

\bibitem{zheng2023judging}
Zheng, L., Chiang, W.L., Sheng, Y., Zhuang, S., Wu, Z., Zhuang, Y., Lin, Z., Li, Z., Li, D., Xing, E., et~al.: Judging llm-as-a-judge with mt-bench and chatbot arena. arXiv preprint arXiv:2306.05685  (2023)

\bibitem{zhu2023ChatMed}
Zhu, W., Wang, X.: Chatmed: A chinese medical large language model. \url{https://github.com/michael-wzhu/ChatMed} (2023)

\end{thebibliography}





% LNCS Homepage, \url{http://www.springer.com/lncs}, last accessed 2023/10/25
% \end{thebibliography}
\end{document}